\def\year{2021}\relax
\documentclass[letterpaper]{article} 
\usepackage{aaai21}  
\usepackage{times}  
\usepackage{helvet} 
\usepackage{courier}  
\usepackage[hyphens]{url}  
\usepackage{graphicx} 
\def\UrlFont{\rm}  
\usepackage{natbib}  
\usepackage{caption} 
\usepackage{cite}

\usepackage{color} 
\usepackage{multirow}
\usepackage[switch]{lineno}  %
\usepackage{booktabs}
\usepackage{arydshln}

\author{

    Fei Yu{\rm}\thanks{indicates equal contribution.},
    Jiji Tang\footnotemark[1],
    Weichong Yin,
    Yu Sun,
    Hao Tian,
    Hua Wu,
    Haifeng Wang
    
}
\affiliations{

    Baidu Inc., Beijing, China\\
    \{yufei07, tangjiji, yinweichong, sunyu02, tianhao, wu\_hua, wanghaifeng\}@baidu.com

}

\title{ERNIE-ViL: Knowledge Enhanced Vision-Language \\
Representations through Scene Graphs}

\begin{document}
\maketitle

\begin{abstract}
We propose a knowledge-enhanced approach, ERNIE-ViL, which incorporates structured knowledge obtained from scene graphs to learn joint representations of vision-language. ERNIE-ViL tries to build the detailed semantic connections (objects, attributes of objects and relationships between objects) across vision and language, which are essential to vision-language cross-modal tasks. Utilizing scene graphs of visual scenes, ERNIE-ViL constructs Scene Graph Prediction tasks, i.e., Object Prediction, Attribute Prediction and Relationship Prediction tasks in the pre-training phase. Specifically, these prediction tasks are implemented by predicting nodes of different types in the scene graph parsed from the sentence. Thus, ERNIE-ViL can learn the joint representations characterizing the alignments of the detailed semantics across vision and language. After pre-training on large scale image-text aligned datasets, we validate the effectiveness of ERNIE-ViL on 5 cross-modal downstream tasks. ERNIE-ViL achieves state-of-the-art performances on all these tasks and ranks the first place on the VCR leaderboard with an absolute improvement of 3.7\%.

\end{abstract}

\section{Introduction}
Motivated by pre-trained models like BERT \cite{devlin2018bert} and GPT \cite{radford2018improving} which have significantly improved the performance of many NLP tasks, researchers \cite{lu2019vilbert,li2019unicoder,su2019vl,li2019visualbert,chen2019uniter} have noticed the importance of pre-training for vision-language tasks, e.g., Visual Question Answering(VQA) \cite{antol2015vqa} and 
Visual Commonsense Reasoning (VCR) \cite{zellers2019recognition}.

\begin{figure*}[t]
\centering
\begin{tabular}{ccc} 
\multicolumn{1}{c}{(a) Objects}  & \multicolumn{1}{c}{(b) Attributes}   & \multicolumn{1}{c}{(c) Relationships}         \\ 
\hline
\begin{minipage}{0.3\textwidth}
\centering
\includegraphics[width=0.72\linewidth]{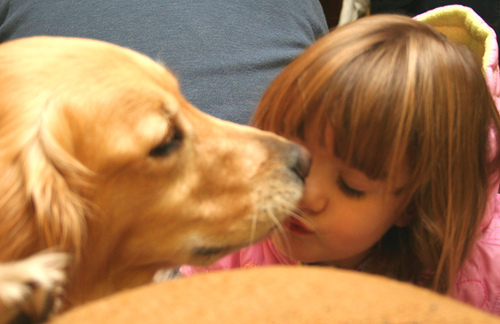}
\end{minipage}
 & 
 \begin{minipage}{0.3\textwidth}
 \centering
\includegraphics[width=0.64\linewidth]{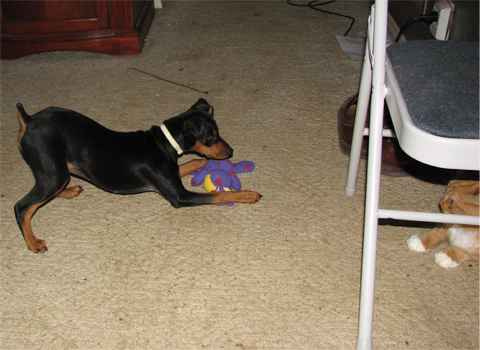}
\end{minipage}
 & 
\begin{minipage}{0.3\textwidth}
\centering
\includegraphics[width=0.76\linewidth]{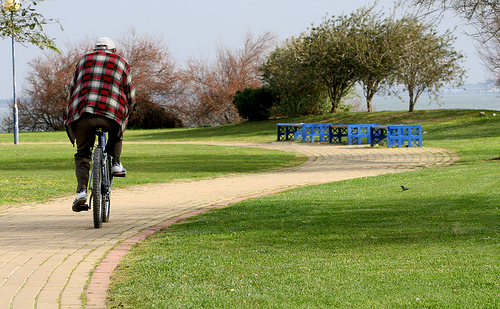}
\end{minipage} 
\\ 
\multicolumn{1}{c}{\small{A tan \textcolor{red}{\textbf{dog}} and a little girl kiss.}}
&
\multicolumn{1}{c}{\small{A black dog playing with a \textcolor{red}{\textbf{purple}} toy.}}
&
\multicolumn{1}{c}{\small{
A man in red plaid \textcolor{red}{\textbf{rides}} his bike in a park.}}
\\

\begin{minipage}{0.3\textwidth}
\centering
\includegraphics[width=0.72\linewidth]{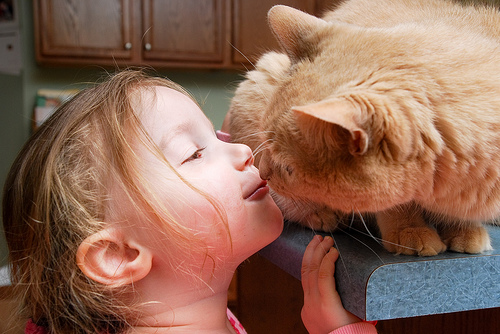}
\end{minipage}
 & 
 \begin{minipage}{0.3\textwidth}
 \centering
\includegraphics[width=0.64\linewidth]{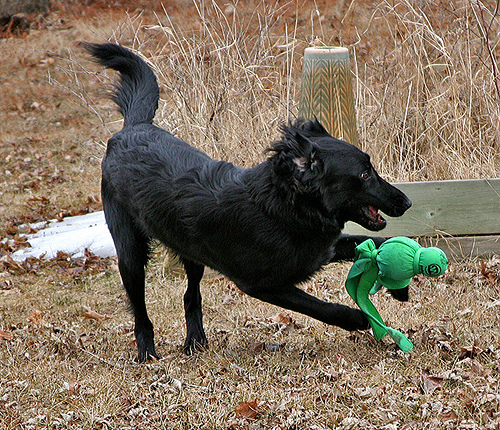}
\end{minipage}
 & 
\begin{minipage}{0.30\textwidth}
\centering
\includegraphics[width=0.80\linewidth]{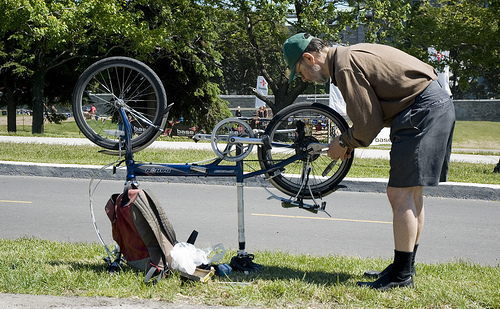}
\end{minipage} 
\\
\multicolumn{1}{c}{\small{The little girl is kissing the brown \textcolor{red}{\textbf{cat}}.}}
&
\multicolumn{1}{c}{\small{A black dog playing with a \textcolor{red}{\textbf{green}} toy.}}
&
\multicolumn{1}{c}{\small{
An older man \textcolor{red}{\textbf{repairing}} a bike tire in a park.}}
\\
\end{tabular}

\caption{Similar scene pairs from the Flick30K datasets \cite{young2014image}. It is the detailed semantics that determine the interpretation of the scenes, objects (dog, cat) in scene pair (a), attributes(purple, green) in scene pair (b) and relationships(rides, repairing) in scene pair (c). }
\label{tab_nodecase}
\end{figure*}

Existing vision-language pre-training methods attempt to learn joint representations through visual grounding tasks on large image-text datasets, including Masked Language Modelling based on randomly-masked sub-words, Masked Region Prediction and Image-Text Matching at the image/text-level. However, based on randomly-masking and predicting the sub-words, current models did not distinguish common words and words describing the detailed semantics \cite{johnson2015image},  e.g., objects(``man", ``boat"), attributes of objects(``boat is white"), relationships between objects(``man standing on boat"). These methods neglect the importance of constructing detailed semantic alignments across vision and language, therefore the trained models can not well represent fine-grained semantics required by some real scenes. As shown in Figure \ref{tab_nodecase}, the detailed semantics are essential to distinguish the listed scenes which mainly differ in objects, attributes and relationships. Hence, better joint vision-language representations should characterize detailed semantic alignments across the modalities.

Inspired by the knowledge masking strategy of ERNIE \cite{sun2019ernie2}, which aims at learning more structured knowledge by masking phrases and named entities rather than individual sub-words, we propose ERNIE-ViL, that incorporates knowledge obtained from scene graphs \cite{johnson2015image} to construct better representations for vision-language joint modelling. Through constructing Scene Graph Prediction tasks, ERNIE-ViL puts more emphasis on detailed semantic alignments across vision and language. Concretely, we implement these pre-training tasks by masking and predicting different types of nodes in the scene graph parsed from the sentence. By concentrating on understanding detailed semantic words rather than common words, these Scene Graph Prediction tasks force the model to extract object/attribute/relationship information from the visual modality, thus establish semantic connections between vision and language.
Pre-training with the Scene Graph Prediction tasks, ERNIE-ViL learns the  vision-language detailed semantic alignments. 

We pre-train ERNIE-ViL on two large commonly-used image-text out-of-domain datasets, namely Conceptual Captions \cite{sharma2018conceptual} and SBU Captions \cite{ordonez2011im2text}. To evaluate the performance of ERNIE-ViL, we conduct experiments on various vision-language tasks,  (1) Visual Question Answering (VQA 2.0) \cite{antol2015vqa}, (2) Visual Commonsense Reasoning (VCR) \cite{zellers2019recognition}, (3) Region-to-Phrase Grounding (RefCOCO+) \cite{kazemzadeh2014referitgame}, (4, 5) Image-text Retrieval / Text-image Retrieval (Flickr30K) \cite{young2014image}. On all these tasks, ERNIE-ViL obtains significant improvements compared to those models pretrained on the same datasets. Especially, on the Region-to-Phrase grounding task that 
relies more   heavily on detailed semantic alignments, we achieve an improvement of  2.4\% on both testsets. To compare with the models pretrained on both out-of-domain and in-domain datasets, we continually pre-train ERNIE-ViL on MS-COCO  \cite{lin2014microsoft}  and Visual-Genome \cite{krishna2017visual} (in-domain datasets for downstream tasks). ERNIE-ViL achieves the  state-of-the-art performances on all downstream tasks. Also ERNIE-ViL obtains the best single model performance and ranks the first place on the leaderboard with an absolute improvement of 3.7\% on the Q$\to$AR task compared to the state-of-the-art performance. And our code and pre-trained models are scheduled to be public.
 
Overall, the contributions of our method are three-folds:
\begin{itemize}
 \item To the best of our knowledge, ERNIE-ViL is the first work that has introduced structured knowledge to enhance vision-language pre-training.
\item ERNIE-ViL constructs Scene Graph Prediction tasks during the pre-training of vision-language joint representations, putting more emphasis on the cross-modal detailed semantics alignments.
\item ERNIE-ViL achieves state-of-the-art performances on 5 downstream cross-modal tasks and ranks the first place on the VCR leaderboard.
\end{itemize}

\begin{figure*}[t]
    \centering
    \includegraphics[width=1.95\columnwidth]{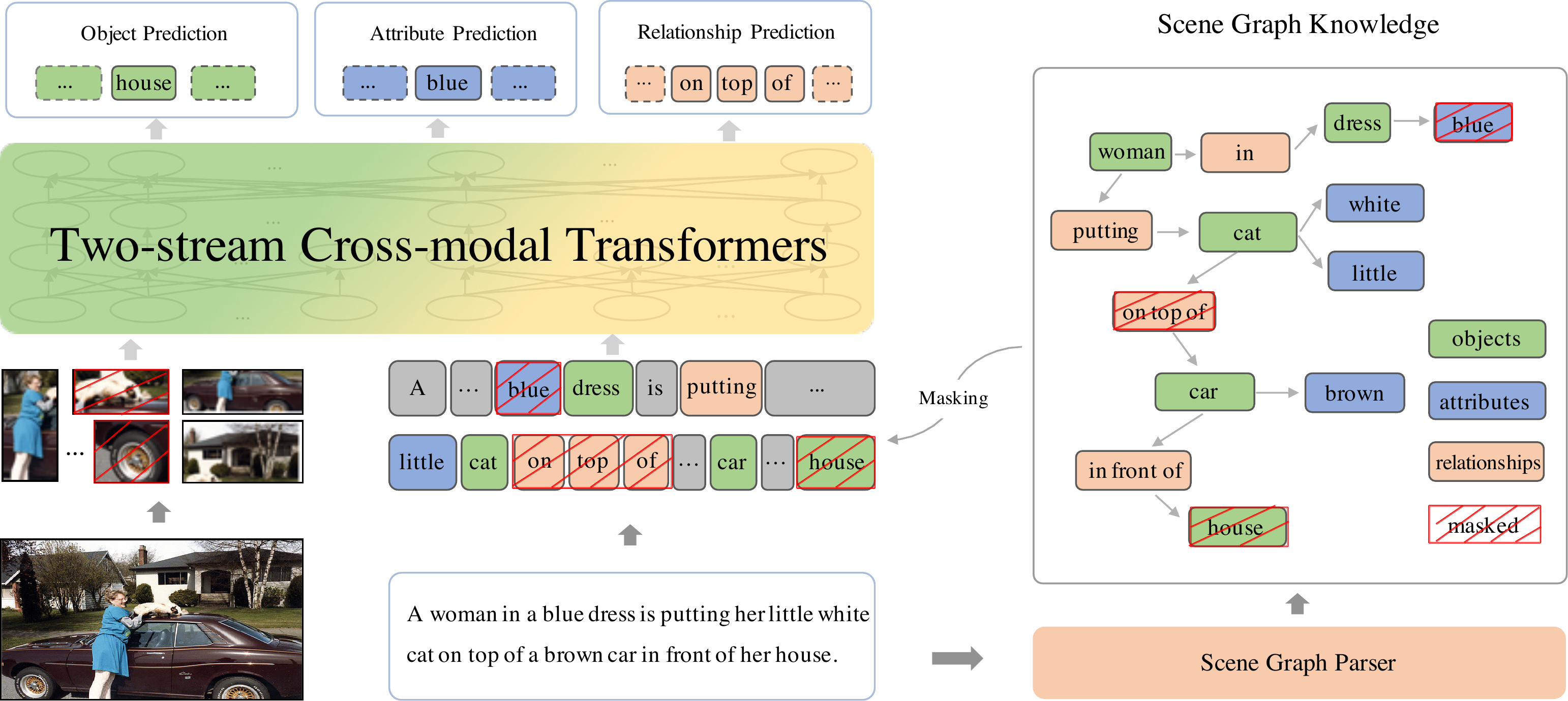}
    \caption{Illustration of Scene Graph Prediction tasks for ERNIE-ViL. Given detected regions of the image and token sequence of the text, ERNIE-ViL uses a two-stream cross-modal Transformers network to model the joint vision-language representations. Based on the scene graph parsed from the text using Scene Graph Parser, we construct Object Prediction, Attribute Prediction and Relationship Prediction tasks to learn cross-modal detailed semantics alignments.}
    \label{fig_framework}
\end{figure*}

\section{Related Works}
\subsection{Cross-modal Pre-training}
Inspired by text pre-training models \cite{devlin2018bert}, many cross-modal pre-training models for vision-language have been proposed.
These researchers put their efforts mainly on three aspects, which are model architecture, pre-training tasks and pre-training data.

\begin{itemize}
 \item \textbf{Model Architecture} Current works are based on different variables of Transformers \cite{vaswani2017attention}. Most of them \cite{li2019unicoder,su2019vl,zhou2019unified,li2019visualbert,huang2020pixel} used a uniform cross-modal Transformer modelling both image and text representations, while the others like ViLBERT \cite{lu2019vilbert} and LXMERT \cite{tan2019lxmert} were based on two-stream cross-modal Transformers, which brings more specific representations for images and texts. 

\item \textbf{Pre-training Tasks} Inspired by the pre-training tasks in text  models, Masked Language Model and similar Masked Region Prediction tasks \cite{lu2019vilbert} are utilized in cross-modal pre-training. And similar to Next-Sentence Prediction, Image-Text Matching \cite{lu2019vilbert,su2019vl,chen2019uniter} task is also widely used. However, based on randomly masking and predicting sub-words, these methods did not distinguish the common words and words describing the detailed semantics. Hence, the  cross-modal fine-grained semantic alignments cannot be well characterized in those learned joint representations.

\item \textbf{Pre-training Data} Unlike text pre-training models that can leverage tremendous natural language data, vision-language tasks require high-quality aligned image-text data that are hard to obtain. Conceptual Captions\cite{sharma2018conceptual} and SBU Captions\cite{ordonez2011im2text} are two widely-used datasets for image-text pre-training, with 3.0M and 1.0M image-description pairs respectively. These two datasets are out-of-domain for vision-language downstream tasks, while some existing works \cite{chen2019uniter,huang2020pixel}  incorpate in-domain datasets, such as MS-COCO  and Visual-Genome, that are highly correlated with downstream tasks.
\end{itemize}

\subsection{Scene Graph}

Scene graphs contain structured knowledge of visual scenes, including the present objects, attributes of objects, and relationships between objects. As a beneficial prior knowledge describing the detailed semantics of images and captions, scene graphs have led to many state-of-the-art models in image captioning \cite{yang2019auto}, image retrieval \cite{wu2019unified}, VQA \cite{zhang2019empirical} and image generation \cite{johnson2018image}.

\section{Approach}
 In this section, we first introduce the architecture of ERNIE-ViL. Then we  illustrate our newly-proposed Scene Graph Prediction tasks. Finally, pre-training with Scene Graph Prediction tasks in ERINE-ViL is introduced.
 
\subsection{Model Architecture}

The vision-language model aims at learning the joint representations that integrates information of both modalities and the alignments across the modalities. The inputs of ERNIE-ViL are a sentence and an image. Given a sequence of words and an image, we introduce the methods to embed the inputs to the feature space and the vision-language encoder.

\paragraph{Sentence Embedding} We adopt the similar word pre-prossessing method as BERT. The input sentence is tokenized into sub-word tokens using WordPiece approach. Special tokens such as $\textit{[CLS]}$ and $\textit{[SEP]}$ are also added to the tokenized text sequence to form the text sequence as $\{\textit{[CLS]}, w1,\dots w_{T} ,\textit{[SEP]}\}$. The final embedding for each sub-word token is generated by combining its original word embedding, segment embedding and sequence position embedding.

\paragraph{Image Embedding} For the image, we first use a pre-trained object detector to detect the salient image regions from the image. The pooling features before multi-class classification layer are utilized as the region features. We also encode the location features for each region via a 5-dimensional vector $(\frac{x_1}{W},\frac{y_1}{H},\frac{x_2}{W},\frac{y_2}{H},\frac{(y_2-y_1)(x_2-x_1)}{WH})$ for the region position and the fraction of image area covered, where $(x_1,y_1)$ and $(x_2,y_2)$ denote
the coordinates of top-left and bottom-right corner while $W$ and $H$ are the width and  height of the input image. The location vectors are projected to form the location features, which are then summed with the region visual features. We also add a special feature $\textit{[IMG]}$ that denotes the representation of the entire image (i.e. mean-pooled visual features with a spatial encoding corresponding to the entire image) to form the final region sequence $\{\textit{[IMG]}, v_{1},\dots, v_{I}\}$. 

\paragraph{Vision-Language Encoder} Given the embedding of image regions and the words for the sentence $\{\textit{[IMG]}, v_{1},\dots, v_{I} , \textit{[CLS]}, w1,\dots w_{T} ; \textit{[SEP]}\}$, we use two-stream cross-modal Transformers to joint model the intra-modal and inter-modal representations. Similar to ViLBERT \cite{lu2019vilbert}, ERNIE-ViL consists of two parallel Transformer encoders for image and text segments, which are cross-attended with cross-modal Transformer blocks. The model outputs embeddings for each input of both the image and text. We take $h_{\textit{[IMG]}}$ and $h_{\textit{[CLS]}}$ as the holistic image and text representations.

\subsection{Scene Graph Prediction}\label{sec_task}


Detailed semantics, includes objects, attributes of objects, and relationships between objects, are essential to the understanding of visual scenes \cite{johnson2015image}. As the scene shown in Figure \ref{fig_framework}, detailed semantics describes the visual scene from different aspects. The objects, such as ``cat", ``car", ``woman" are the fundamental elements in the scene. And associated attributes, such as ``little", ``brown", ``blue"  characterize shape and color of objects. Relationships such as ``on top of", ``putting" represent the spatial connections and actions between objects. Therefore detailed semantics are crucial in accurately understanding visual scenes. 
Since the goal of vision-language joint representations is to engrave the semantic connections across modalities, detailed semantic alignments are significantly important in cross-modal learning.

Scene graphs encode various fine-grained semantic information. Utilizing structured knowledge obtained from scene graphs, ERNIE-ViL learns the cross-modal detailed semantic alignments. As shown in Figure \ref{fig_framework}, according to the scene graph parsed from the text, we construct the corresponding Scene Graph Prediction tasks, including Object Prediction task, Attribute Prediction task, and Relationship Prediction task. These tasks force ERNIE-ViL to model the correlations of detailed semantics across modalities. For example, as the relationship words ``on top of" is masked, based on the language context, the model may predict that the missing word is ``under" or ``into". These words are grammatically fluent in the sentence, but are inconsistent with the scene ``the cat is on top of the car". Through training the Relationship Prediction task, the model obtains the spatial relation of the corresponding objects(``car", ``cat") from the image, thus can accurately predict that the missing word is ``on top of".  Through constructing Scene Graph Prediction tasks,  ERNIE-ViL learns cross-modal detailed semantic alignments.

\paragraph{Scene graph parsing}
Given the text sentence $\mathbf{w}$,  we parse it into a scene graph\cite{johnson2015image}, which denotes as  $G(\mathbf{w})=<O(\mathbf{w}),E(\mathbf{w}),K(\mathbf{w})>$, where $O(\mathbf{w})$ is the set of objects mentioned in $\mathbf{w}$, $E(\mathbf{w})\subseteq O(\mathbf{w})\times R(\mathbf{w})\times O(\mathbf{w})$ is the set of hyper-edges representing relationship triplets, and $R(\mathbf{w})$ is the set of relationship nodes between object nodes. $K(\mathbf{w})\subseteq O(\mathbf{w})\times A(\mathbf{w})$ is the set of attribute pairs, where  $A(\mathbf{w})$ is the set of attribute nodes associated with object nodes. Scene graphs describe the objects in more details with various associated attributes  and relationships between objects. Thus integrating the knowledge  of scene graphs can benefit learning  more fine-grained joint representations for the vision-language. In this paper, the Scene Graph Parser provided by Anderson \cite{anderson2016spice} is adopted to parse texts to scene graphs. For a more intuitive understanding, we illustrate a specific case for the parsed scene graph  from the text in  
Table \ref{tab_sg}.

\begin{table}
\begin{center}
\begin{tabular}{c|c}
\hline
\multicolumn{1}{c|}{\multirow{3}{*}{sentence: $\mathbf{w}$}} & A woman in blue dress is putting  \\ 
\multicolumn{1}{c|}{}  &  her little white cat on top of a \\
\multicolumn{1}{c|}{}  &  brown car in front of her house. \\ \hline

\multicolumn{1}{c|}{\multirow{1}{*}{objects:$O(\mathbf{w})$}} & dress, woman, cat, car, house \\ \hline

\multicolumn{1}{c|}{\multirow{1}{*}{relationships:$R(\mathbf{w})$}} & in, putting, on-top-of, in-front-of \\ \hline

\multicolumn{1}{c|}{\multirow{1}{*}{ attributes: $A(\mathbf{w})$}} & blue, white, little, brown  \\ \hline

\end{tabular}
\end{center}
\caption{The scene graph parsed from the caption of the visual scene. For simplicity, we only list all the nodes leaving out the connections between them. }

\label{tab_sg}
\end{table}

\subsubsection{Object Prediction}

Objects are the dominant elements of  visual scenes, thus playing an important role in constructing the representations of semantic information. Predicting the objects forces the model to build the vision-language connections at object level. 

Firstly, for all the object nodes in the scene graph, we randomly select 30\% of them to mask. And for each selected object node $O(\mathbf{w})$, we replace it with the special token $\textit{[MASK]}$ in probability of 80\%, another random token in probability of 10\%, and keep it in probability of 10\%. Note that the objects actually correspond to the sub-sequences of text in the sentence, therefore the object masking are implemented by masking the corresponding sub-sequences in the text.

For Object Prediction, ERNIE-ViL recover these masked object tokens, denoted as $\mathbf{w}_{o_{i}}$, based on their surrounding words $\mathbf{w}$ and all image regions $\mathbf{v}$, by minimizing the negative log-likelihood:
\begin{equation}
    \mathcal{L}_{obj}(\theta)=-E_{(\mathbf{w},\mathbf{v})\sim D}\log(P(\mathbf{w}_{o_{i}}| \mathbf{w}_{\setminus \mathbf{w}_{o_{i}}},\mathbf{v}))
\end{equation}


\subsubsection{Attribute Prediction}
Attributes characterize the specific information of the visual objects, such as color or shape of the objects, therefore representing the detailed information in the visual scenes in more fine-grained level.

Similarly, we randomly select 30\% of the attribute pairs in the scene graph, and the mask strategy here is the same as that in Object Prediction. Since the attribute nodes in the scene graph are attached to objects, we keep the associated object while masking out the attribute node $A(\mathbf{w})$ in each selected $K(\mathbf{w})\subseteq O(\mathbf{w})\times A(\mathbf{w})$.

Given object words $w_{o_{i}}$ in attribute pair $ \langle w_{o_i},w_{a_{i}} \rangle$, Attribute Prediction is to recover the masked tokens $w_{a_{i}}$ of attribute pairs. Based on the object tokens $w_{o_{i}}$ , other surrounding words $\mathbf{w}$ and all image regions v, Attribute Prediction minimizes the negative log-likelihood:
\begin{equation}
    \mathcal{L}_{attr}(\theta)=-E_{(\mathbf{w},\mathbf{v})\sim D}\log(P(\mathbf{w}_{a_{i}}|\mathbf{w}_{o_{i}}, \mathbf{w}_{\setminus\mathbf{w}_{a_{i}}},\mathbf{v}))
\end{equation}


\subsubsection{Relationship Prediction}

Relationships describe the actions (semantic) or  relative position (geometry) between the objects of the visual scenes, which contributes to distinguish scenes with same objects but different relationships.

Thus, ERNIE-ViL constructs the Relationship Prediction task to learn cross-modal  relationships connections. When performing the mask strategy of selected relationship triplets $E(\mathbf{w})\subseteq O(\mathbf{w})\times R(\mathbf{w}) \times O(\mathbf{w})$, we keep the objects and mask out the relationship node $R(\mathbf{w})$. Specifically,  
 given object tokens $\mathbf{w}_{o_{i1}},\mathbf{w}_{o_{i2}}$ in relationship triplet $ \langle \mathbf{w}_{o_{i1}},\mathbf{w}_{r_i},\mathbf{w}_{o_{i2}} \rangle$, this task recovers the masked relationship tokens, predicting the probability for each masked relation tokens $\mathbf{w}_{r_{i}}$. Thus the context for the prediction is  the given object tokens $\mathbf{w}_{o_{i1}},\mathbf{w}_{o_{i2}}$, other surrounding words from the text and all image regions $\mathbf{v}$. The loss for this task is:

\begin{equation}
    \mathcal{L}_{rel}(\theta)=-E_{(\mathbf{w},\mathbf{v})\sim D}\log(P(\mathbf{w}_{r_{i}}|\mathbf{w}_{o_{i1}},\mathbf{w}_{o_{i2}}, \mathbf{w}_{\setminus \mathbf{w}_{r_i}},\mathbf{v}))
\end{equation}

\subsection{Pre-training with Scene Graph Prediction}

Simliar to ViLBERT\cite{lu2019vilbert}, ERNIE-ViL also adopts Masked Language Modelling(MLM) to capture the syntactic and lexical information in the text. Moreover, Masked Region Prediction and Image-text Matching are utilized for visual modality and cross-modality respectively. The losses for all these pre-training tasks are summed.

\section{Experiments}
\subsection{Training ERNIE-ViL}

\paragraph{Pre-training Data} We use the Conceptual Captions (CC) dataset \cite{sharma2018conceptual} and SBU Captions (SBU) dataset \cite{ordonez2011im2text} as pre-training data. CC is a collection of 3.3 million image-caption pairs automatically scraped from alt-text enabled web images and SBU is a similar vision-language dataset which has 1.0 million image-caption pairs. Since some links have become broken, only about 3.0 million pairs for CC dataset and 0.8 million pairs for SBU dataset are available and utilized in our experiments. Note that CC and SBU are image-caption pairs automatically collected from the web and have no intersections with the down-stream task datasets, thus act as out-of-domain datasets for training vision-language models.


\paragraph{Implementation Details}  
For each image-text pair in the training, the pre-processing is performed as follows. For the image, we adopt Faster R-CNN \cite{anderson2018bottom} to select salient image regions and extract region features. Specifically, regions with class detection probability exceeds a confidence threshold of 0.2 are selected and 10 to 36 boxes  are kept. And for each kept region, the mean-pooled convolutional representation  is used as the region feature. For the text, we parse the scene graph from the sentence using the Scene Graph Parser and adopt WordPieces to tokenize the sentence similar to BERT.

For the masking strategies, we randomly mask 15\% of tokens, 30\% of scene graph nodes, and 15\% of image regions. For the Image-text Matching task, we randomly select a image for each text to form the negative image-text pair. Note that only items in the positive pairs will be considered for token and region prediction tasks.

We train ERNIE-ViL on two scale settings: ERNIE-ViL-base and ERNIE-ViL-large, which mainly differ in model depth of the text stream. The detailed settings of text and visual streams are shown in Table \ref{tab_setting}. And similar to VilBERT\cite{lu2019vilbert}, cross-transformers are used to co-at tent the two streams. We initialize the text stream parameters from ERNIE 2.0 \cite{sun2019ernie2}, and implement ERNIE-ViL via PaddlePaddle. After then, ERINE-ViL is pre-trained on a total batch size of 512 for  700k steps on 8 V100 GPUs, using adam optimizer with initial learning rates of 1e-4 and Noam \cite{vaswani2017attention} as learning rate decay schedule.

\begin{table}[t]
\centering
\setlength{\tabcolsep}{1.8mm}{
\small
 \begin{tabular}{c|cccc|cccc}
 \hline
& \multicolumn{4}{c|}{Base} & \multicolumn{4}{c}{Large}  \\
 &$L$ & $H$ & $A$ & $F$ & $L$ & $H$ & $A$ & $F$ \\ \hline
Text   &12 &768 &12 &3072 &24 &1024 &16 & 4096   \\
Visual  & 6 &1024& 8 & 1024 &6 &1024 &16 &4096  \\
\hline
\end{tabular}
}
\caption{Settings for ERNIE-ViL model. $L$: number of layers, $H$ : hidden size, $A$ : number of self-attention heads, $F$ : feed-forward/filter size.}
\label{tab_setting}

\end{table}

\subsection{Downstream Tasks}
\subsubsection{Visual Commonsense Reasoning (VCR)}

The Visual Commonsense Reasoning (VCR) \cite{zellers2019recognition} task contains two sub-tasks: visual question answering (Q$\to$A) and answer justification (QA$\to$R), which are  both multiple choice problems. The holistic setting (Q$\to$AR) requires both the chosen answer and chosen rationale to be correct. In visual question answering (Q$\to$A) task, we concatenate the question and each candidate answer for the language modality.  We take dot product of final hidden state  $h_{[\textit{CLS}]}$  and $h_{[\textit{IMG}]}$ to predict matching score with an additional FC layer. For the answer justification (QA$\to$R) task, we concatenate the question, the answer and each candidate rationale as the input of the text stream. Similar with UNITER \cite{chen2019uniter}, a second-stage pre-training is adopted on VCR dataset. And then we fine-tune the model over 6 epochs with a batch size of 64 and adopt Adam optimizer with initial learning rate of 1e-4.

\subsubsection{Visual Question Answering (VQA)}

The VQA task requires answering natural language questions according to images. VQA 2.0 dataset \cite{antol2015vqa} contains  204k images and 1.1M questions about these images. Also additional question-answer pairs from Visual Genome are used for data augmentation as in UNITER \cite{chen2019uniter}.
We treat VQA as a multi-label classification task – assigning a soft target score to each answer based on its relevancy to the 10 human answer responses. We take dot product of final hidden state  $h_{[\textit{CLS}]}$  and $h_{[\textit{IMG}]}$ to map this representation into 3,129 possible answers with an additional two-layer MLP.  
Fine-tuning of VQA model is performed over 12 epochs on batch size of 256 and using Adam optimizer with initial learning rate of 1e-4.

\begin{table*}[t]
  \centering
  \renewcommand{\arraystretch}{1.2}
  \tabcolsep 6.4pt
   \small
 \begin{tabular}{llllllllllll}

\toprule[0.7pt]
\multicolumn{1}{c}{\multirow{2}{*}{Domains}} & \multicolumn{1}{c}{\multirow{2}{*}{Models}} & 
&
\multicolumn{3}{c}{VCR} &
&
\multicolumn{3}{c}{RefCOCO+} \\
\cline{8-10}
\cline{4-6}
 \multicolumn{1}{c}{} & \multicolumn{1}{c}{} & 
 \multicolumn{1}{c}{} & 

 \multicolumn{1}{c}{\texttt{Q$\to$A}} & \multicolumn{1}{c}{\texttt{QA$\to$R}} & \multicolumn{1}{c}{\texttt{Q$\to$AR}} &
 \multicolumn{1}{c}{} & 
 \multicolumn{1}{c}{\texttt{val}} &
 \multicolumn{1}{c}{\texttt{testA}} & \multicolumn{1}{c}{\texttt{testB}} \\
\hline

\multirow{7}{*}{Out-of-domain} & \multicolumn{1}{c}{UNITER-base} & \multicolumn{1}{c}{} & \multicolumn{1}{c}{-} & \multicolumn{1}{c}{-} & \multicolumn{1}{c}{-} & \multicolumn{1}{c}{} & \multicolumn{1}{c}{72.78} & \multicolumn{1}{c}{-} & \multicolumn{1}{c}{-} \\

& \multicolumn{1}{c}{Unicoder-VL-base} & \multicolumn{1}{c}{} & \multicolumn{1}{c}{72.6(73.4)} & \multicolumn{1}{c}{74.5(74.4)} & \multicolumn{1}{c}{54.4(54.9)} & \multicolumn{1}{c}{} & \multicolumn{1}{c}{-} & \multicolumn{1}{c}{-} & \multicolumn{1}{c}{-} \\

& \multicolumn{1}{c}{ViLBERT-base} & \multicolumn{1}{c}{} & \multicolumn{1}{c}{72.42(73.3)} & \multicolumn{1}{c}{74.47(74.6)} & \multicolumn{1}{c}{54.04(54.8)} & \multicolumn{1}{c}{} & \multicolumn{1}{c}{72.34} & \multicolumn{1}{c}{78.52} & \multicolumn{1}{c}{62.61} \\

& \multicolumn{1}{c}{VLBERT-base} & \multicolumn{1}{c}{} & \multicolumn{1}{c}{73.8(-)} & \multicolumn{1}{c}{74.4(-)} & \multicolumn{1}{c}{55.2(-)} & \multicolumn{1}{c}{} & \multicolumn{1}{c}{71.60} & \multicolumn{1}{c}{77.72} & \multicolumn{1}{c}{60.99} \\

& \multicolumn{1}{c}{ERNIE-ViL-base} & \multicolumn{1}{c}{} & \multicolumn{1}{c}{76.37(77.0)} & \multicolumn{1}{c}{79.65(80.3)} & \multicolumn{1}{c}{61.24(62.1)} & \multicolumn{1}{c}{} & \multicolumn{1}{c}{74.02} & \multicolumn{1}{c}{80.33} & \multicolumn{1}{c}{\textbf{64.74}} \\

\cline{2-10}

& \multicolumn{1}{c}{VLBERT-Large} & \multicolumn{1}{c}{} & \multicolumn{1}{c}{75.5(75.8)} & \multicolumn{1}{c}{77.9(78.4)} & \multicolumn{1}{c}{58.9(59.7)} & \multicolumn{1}{c}{} & \multicolumn{1}{c}{72.59} & \multicolumn{1}{c}{78.57} & \multicolumn{1}{c}{62.30} \\

& \multicolumn{1}{c}{ERNIE-ViL-Large} & \multicolumn{1}{c}{} & \multicolumn{1}{c}{\textbf{78.52(79.2)}} & \multicolumn{1}{c}{\textbf{83.37(83.5)}} & \multicolumn{1}{c}{\textbf{65.81(66.3)}} & \multicolumn{1}{c}{} & \multicolumn{1}{c}{\textbf{74.24}} & \multicolumn{1}{c}{\textbf{80.97}} & \multicolumn{1}{c}{64.70} \\
\hline

\multirow{3}{*}{\shortstack{Out-of-domain\\+ in-domain}} & \multicolumn{1}{c}{UNITER-large} & \multicolumn{1}{c}{} & \multicolumn{1}{c}{77.22(77.3)} & \multicolumn{1}{c}{80.49(80.8)} & \multicolumn{1}{c}{62.59(62.8)} & \multicolumn{1}{c}{} & \multicolumn{1}{c}{75.90} & \multicolumn{1}{c}{81.45} & \multicolumn{1}{c}{66.70}\\

&\multicolumn{1}{c}{VILLA-large} & \multicolumn{1}{c}{} & \multicolumn{1}{c}{78.45(78.9)} & \multicolumn{1}{c}{82.57(82.8)} & \multicolumn{1}{c}{65.18(65.7)} & \multicolumn{1}{c}{} & \multicolumn{1}{c}{\textbf{76.17}} & \multicolumn{1}{c}{81.54} & \multicolumn{1}{c}{66.84}\\

&\multicolumn{1}{c}{ERNIE-ViL-large} & \multicolumn{1}{c}{} & \multicolumn{1}{c}{\textbf{78.98(-)}} & \multicolumn{1}{c}{\textbf{83.70(-)}} & \multicolumn{1}{c}{\textbf{66.44(-)}} & \multicolumn{1}{c}{} & \multicolumn{1}{c}{75.89} & \multicolumn{1}{c}{\textbf{82.37}} & \multicolumn{1}{c}{\textbf{66.91}}\\
\bottomrule[0.7pt]

\end{tabular}

 \begin{tabular}{llllllllllllll}

\toprule[0.7pt]

\multicolumn{1}{c}{\multirow{2}{*}{Domains}} & \multicolumn{1}{c}{\multirow{2}{*}{Models}} &  &
\multicolumn{2}{c}{VQA} &
&
\multicolumn{3}{c}{IR-Flickr30K}
& &
\multicolumn{3}{c}{TR-Flickr30K}
\\
\cline{7-9}
\cline{4-5}
\cline{11-13}
 \multicolumn{1}{c}{} & \multicolumn{1}{c}{} & 
 \multicolumn{1}{c}{} & 

 \multicolumn{1}{c}{\texttt{test-dev}} & 
 \multicolumn{1}{c}{\texttt{test-std}} & 
 \multicolumn{1}{c}{} & 
 \multicolumn{1}{c}{\texttt{R@1}} &
 \multicolumn{1}{c}{\texttt{R@5}} & \multicolumn{1}{c}{\texttt{R@10}} &
 \multicolumn{1}{c}{} & 
 \multicolumn{1}{c}{\texttt{R@1}} &
 \multicolumn{1}{c}{\texttt{R@5}} & \multicolumn{1}{c}{\texttt{R@10}} \\
\hline

\multirow{7}{*}{Out-of-domain} & \multicolumn{1}{c}{UNITER-base} & \multicolumn{1}{c}{} & \multicolumn{1}{c}{71.56} & \multicolumn{1}{c}{-} & \multicolumn{1}{c}{} & \multicolumn{1}{c}{-} & \multicolumn{1}{c}{-} & \multicolumn{1}{c}{-} &
\multicolumn{1}{c}{} &
\multicolumn{1}{c}{-} & \multicolumn{1}{c}{-} & 
\multicolumn{1}{c}{-} \\

& \multicolumn{1}{c}{Unicoder-VL-base} & \multicolumn{1}{c}{} & \multicolumn{1}{c}{-} & \multicolumn{1}{c}{-} & \multicolumn{1}{c}{} & \multicolumn{1}{c}{71.50} & \multicolumn{1}{c}{90.90} & \multicolumn{1}{c}{94.90} & \multicolumn{1}{c}{} &
 \multicolumn{1}{c}{86.20} & \multicolumn{1}{c}{96.30} & \multicolumn{1}{c}{99.00} \\

& \multicolumn{1}{c}{VLBERT-base} & \multicolumn{1}{c}{} & \multicolumn{1}{c}{71.16} & \multicolumn{1}{c}{-} & \multicolumn{1}{c}{} & \multicolumn{1}{c}{-} & \multicolumn{1}{c}{-} & \multicolumn{1}{c}{-} & \multicolumn{1}{c}{} &
 \multicolumn{1}{c}{-} & \multicolumn{1}{c}{-} & \multicolumn{1}{c}{-} \\ 

& \multicolumn{1}{c}{ViLBERT-base} & \multicolumn{1}{c}{} & \multicolumn{1}{c}{70.55} & \multicolumn{1}{c}{70.92} & \multicolumn{1}{c}{} & \multicolumn{1}{c}{58.20} & \multicolumn{1}{c}{84.90} & \multicolumn{1}{c}{91.52} & \multicolumn{1}{c}{} &
 \multicolumn{1}{c}{-} & \multicolumn{1}{c}{-} & \multicolumn{1}{c}{-} \\
 
& \multicolumn{1}{c}{ERNIE-ViL-base} & \multicolumn{1}{c}{} & \multicolumn{1}{c}{73.18} & \multicolumn{1}{c}{73.36} & \multicolumn{1}{c}{} & \multicolumn{1}{c}{74.44} & \multicolumn{1}{c}{92.72} & \multicolumn{1}{c}{95.94} & \multicolumn{1}{c}{} &
 \multicolumn{1}{c}{86.70} & \multicolumn{1}{c}{97.80} & \multicolumn{1}{c}{99.00} \\ 

\cline{2-13}
 & \multicolumn{1}{c}{VLBERT-large} & \multicolumn{1}{c}{} & \multicolumn{1}{c}{71.79} & \multicolumn{1}{c}{72.22} & \multicolumn{1}{c}{} & \multicolumn{1}{c}{-} & \multicolumn{1}{c}{-} & \multicolumn{1}{c}{-} & \multicolumn{1}{c}{} &
 \multicolumn{1}{c}{-} & \multicolumn{1}{c}{-} & \multicolumn{1}{c}{-} \\ 
 & \multicolumn{1}{c}{ERNIE-ViL-large} & \multicolumn{1}{c}{} & \multicolumn{1}{c}{\textbf{73.78}} & \multicolumn{1}{c}{\textbf{73.96}} & \multicolumn{1}{c}{} & \multicolumn{1}{c}{\textbf{75.10}} & \multicolumn{1}{c}{\textbf{93.42}} & \multicolumn{1}{c}{\textbf{96.26}} & \multicolumn{1}{c}{} &
 \multicolumn{1}{c}{\textbf{88.70}} & \multicolumn{1}{c}{\textbf{97.30}} & \multicolumn{1}{c}{\textbf{99.10}} \\ 
 \hline
 \multirow{4}{*}{\shortstack{Out-of-domain\\+ in-domain}} & \multicolumn{1}{c}{UNITER-large} & \multicolumn{1}{c}{} & \multicolumn{1}{c}{73.82} & \multicolumn{1}{c}{74.02} & \multicolumn{1}{c}{} & \multicolumn{1}{c}{75.56} & \multicolumn{1}{c}{94.08} & \multicolumn{1}{c}{96.76} & \multicolumn{1}{c}{} &
 \multicolumn{1}{c}{87.30} & \multicolumn{1}{c}{98.00} & \multicolumn{1}{c}{\textbf{99.20}} \\
 
 & \multicolumn{1}{c}{OSCAR-large} & \multicolumn{1}{c}{} & \multicolumn{1}{c}{73.61} & \multicolumn{1}{c}{73.82} & \multicolumn{1}{c}{} & \multicolumn{1}{c}{-} & \multicolumn{1}{c}{-} & \multicolumn{1}{c}{-} & \multicolumn{1}{c}{} &
 \multicolumn{1}{c}{-} & \multicolumn{1}{c}{-} & \multicolumn{1}{c}{-} \\
 
 & \multicolumn{1}{c}{VILLA-large} & \multicolumn{1}{c}{} & \multicolumn{1}{c}{74.69} & \multicolumn{1}{c}{74.87} & \multicolumn{1}{c}{} & \multicolumn{1}{c}{76.26} & \multicolumn{1}{c}{\textbf{94.24}} & \multicolumn{1}{c}{\textbf{96.84}} & \multicolumn{1}{c}{} &
 \multicolumn{1}{c}{87.90} & \multicolumn{1}{c}{97.50} & \multicolumn{1}{c}{98.80} \\

  & \multicolumn{1}{c}{ERNIE-ViL-large} & \multicolumn{1}{c}{} & \multicolumn{1}{c}{\textbf{74.95}} & \multicolumn{1}{c}{\textbf{75.10}} & \multicolumn{1}{c}{} & \multicolumn{1}{c}{\textbf{76.66}} & \multicolumn{1}{c}{94.16} & \multicolumn{1}{c}{96.76} & \multicolumn{1}{c}{} &
 \multicolumn{1}{c}{\textbf{89.20}} & \multicolumn{1}{c}{\textbf{98.50}} & \multicolumn{1}{c}{\textbf{99.20}} \\
 
\bottomrule[0.7pt]
\hline

\end{tabular}
\caption{Results of downstream vision-language tasks for ERNIE-ViL model, compared with previous state-of-the-art pre-trained models. 
IR: Image Retrieval. TR: Text Retrieval. For VCR task which has private test set, we only report the test results (in parentheses) for ERNIE-ViL models pre-trained on out-of-domain datasets.}
\label{tab_result}
\end{table*}

\subsubsection{Grounding Referring Expressions}

The referring expression task is to localize an image region given a natural language reference. We evaluate the task on RefCOCO+ dataset \cite{kazemzadeh2014referitgame}. Bounding box proposals provided by Mattnet \cite{yu2018mattnet}  are utilized. The representation for each region is denoted by its final hidden state $h_{v_i}$ with an additional FC layer. Each region $i$ is labelled as positive only when the IoU between it and the ground truth box is over 0.5.  We fine-tune the model over 20 epochs with a batch size of 256 and  adopt Adam optimizer with initial learning rate of 1e-4. 

\subsubsection{Image Retrieval \& Text Retrieval}

Caption-based image retrieval is a task of identifying an image from a pool based on a caption describing its content. Flickr30K \cite{young2014image} contains 31,000 images and 5 captions for each image. Adopting the same split in ViLBERT \cite{lu2019vilbert}, we use each of 1,000 images for validation and for testing and the rest for training. We take dot product of final hidden state of $h_{[\textit{CLS}]}$  and $h_{[\textit{IMG}]}$ to predict matching score $s(\mathbf{w},\mathbf{v})$ for each image-text pair with an additional FC layer. We utilize circle loss \cite{sun2020circle} with 20 random negative samples for each image-text pair.
We trained 40 epochs using Adam optimizer with a initial learning rate 1e-5.

\subsection{Results}

\begin{table*}[t]
  \centering
  \renewcommand{\arraystretch}{1.2}
  \tabcolsep 6.4pt
   \small
 \begin{tabular}{llllllllllllllll}

\toprule[0.7pt]

\multirow{2}{*}{\shortstack{initialized text \\ stream parameters }}&

\multicolumn{1}{c}{\multirow{2}{*}{pre-training tasks}}&
&
\multicolumn{1}{c}{VCR} &
&
\multicolumn{1}{c}{VQA} &
&
\multicolumn{1}{c}{RefCOCO+} &
&
\multicolumn{1}{c}{IR} &
&
\multicolumn{1}{c}{TR} &
\\
\cline{4-4}
\cline{6-6}
\cline{8-8}
\cline{10-10}
\cline{12-12}

 & \multicolumn{1}{c}{} &

 &\multicolumn{1}{c}{\texttt{Q$\to$AR(dev)}} &
 &
 \multicolumn{1}{c}{\texttt{dev}} &
 &
 \multicolumn{1}{c}{\texttt{val}} &
 &
 \multicolumn{1}{c}{\texttt{R@1(dev)}} &
 &
  \multicolumn{1}{c}{\texttt{R@1(dev)}} &
 \\
\hline

\multicolumn{1}{c}{BERT} & 

 \multicolumn{1}{c}{w/o \textbf{SGP}} &   \multicolumn{1}{c}{} &
\multicolumn{1}{c}{59.06} &
\multicolumn{1}{c}{} &
\multicolumn{1}{c}{72.38}&
\multicolumn{1}{c}{}&
\multicolumn{1}{c}{72.81} &
\multicolumn{1}{c}{} &
\multicolumn{1}{c}{70.74} &
\multicolumn{1}{c}{} &
\multicolumn{1}{c}{85.00}\\

\multicolumn{1}{c}{BERT} & 

 \multicolumn{1}{c}{w/ \textbf{SGP}} &   \multicolumn{1}{c}{} &
\multicolumn{1}{c}{59.92} &
\multicolumn{1}{c}{} &
\multicolumn{1}{c}{73.04}&
\multicolumn{1}{c}{}&
\multicolumn{1}{c}{73.50} &
\multicolumn{1}{c}{} &
\multicolumn{1}{c}{72.96} &
\multicolumn{1}{c}{} &
\multicolumn{1}{c}{87.40}\\

\multicolumn{1}{c}{ERNIE-2.0} & 

 \multicolumn{1}{c}{w/ \textbf{SGP}} &   \multicolumn{1}{c}{} &
\multicolumn{1}{c}{61.24} &
\multicolumn{1}{c}{} &
\multicolumn{1}{c}{73.18}&
\multicolumn{1}{c}{}&
\multicolumn{1}{c}{74.02} &
\multicolumn{1}{c}{} &
\multicolumn{1}{c}{73.58} &
\multicolumn{1}{c}{} &
\multicolumn{1}{c}{87.80}\\

\bottomrule[0.7pt]
\end{tabular}
\caption{Results of downstream vision-language tasks for ERNIE-ViL pre-trainging with/without Scene Graph Prediction (\textbf{SGP}) tasks, and using different text stream parameters initialization.  
IR \& TR: image retrieval \& text  retrieval on Flickr30K. }
\label{tab_abl}

\end{table*}

\begin{table*}[tb]
\centering
\begin{tabular}{ccccc} 
\toprule[0.7pt]
&Image    & Text  & with \textbf{SGP} task  &   without \textbf{SGP} task      \\  
\hline
1&\begin{minipage}{0.20\textwidth}
\includegraphics[width=0.70\linewidth]{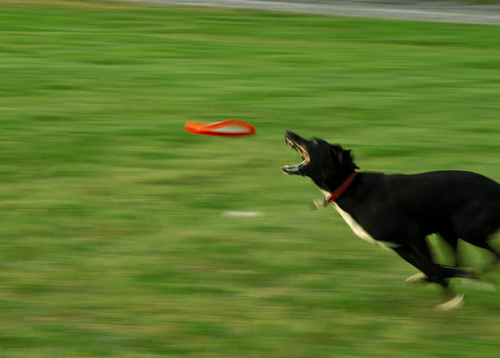}
\end{minipage} 
&
 \begin{minipage}{0.20\textwidth}
\small{ a black dog about to catch a flying \textcolor{red}{\textbf{disc}} .}  
\end{minipage}
& 
\begin{minipage}{0.20\textwidth}
\includegraphics[width=0.7\linewidth]{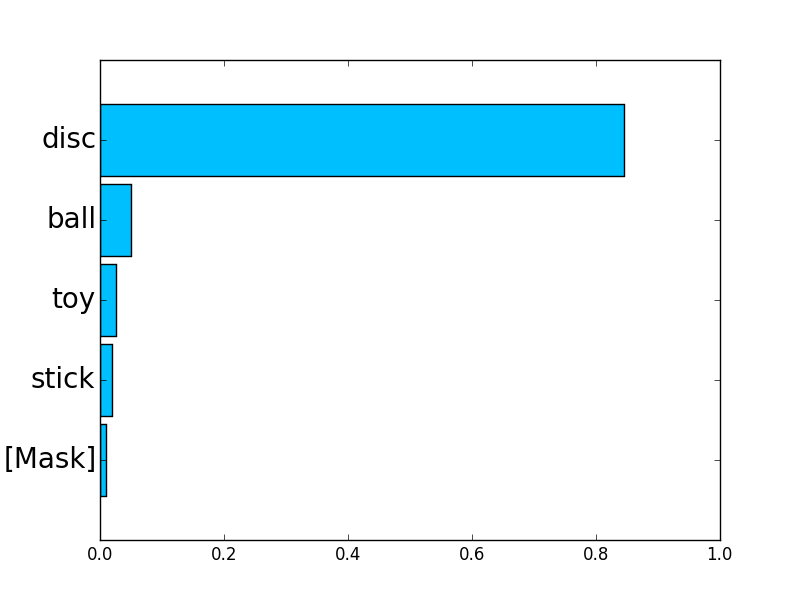}
\end{minipage} 
& 
\begin{minipage}{0.20\textwidth}
\includegraphics[width=0.7\linewidth]{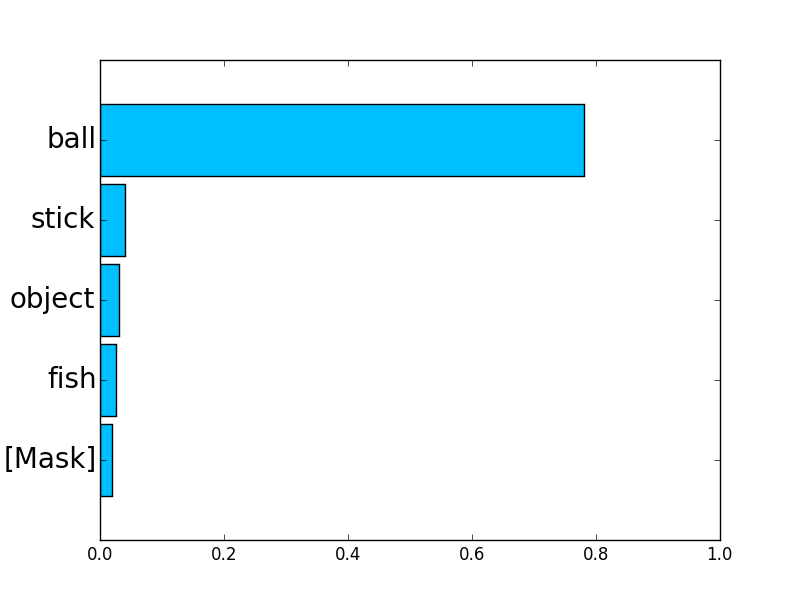}
\end{minipage}
\\

2&\begin{minipage}{0.20\textwidth}
\includegraphics[width=0.70\linewidth]{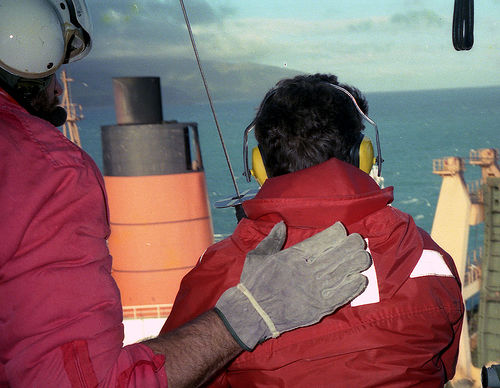}
\end{minipage} 
&
 \begin{minipage}{0.20\textwidth}
\small{two men wearing \textcolor{red}{\textbf{red}} jackets are looking out over some water and one man has yellow earphones on his ears .}  
\end{minipage}
& 
\begin{minipage}{0.20\textwidth}
\includegraphics[width=0.7\linewidth]{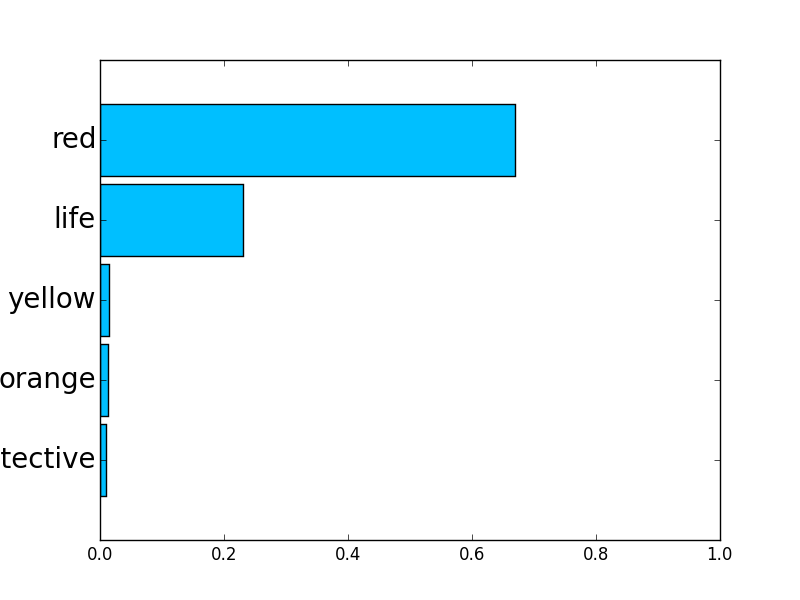}
\end{minipage} 
& 
\begin{minipage}{0.20\textwidth}
\includegraphics[width=0.7\linewidth]{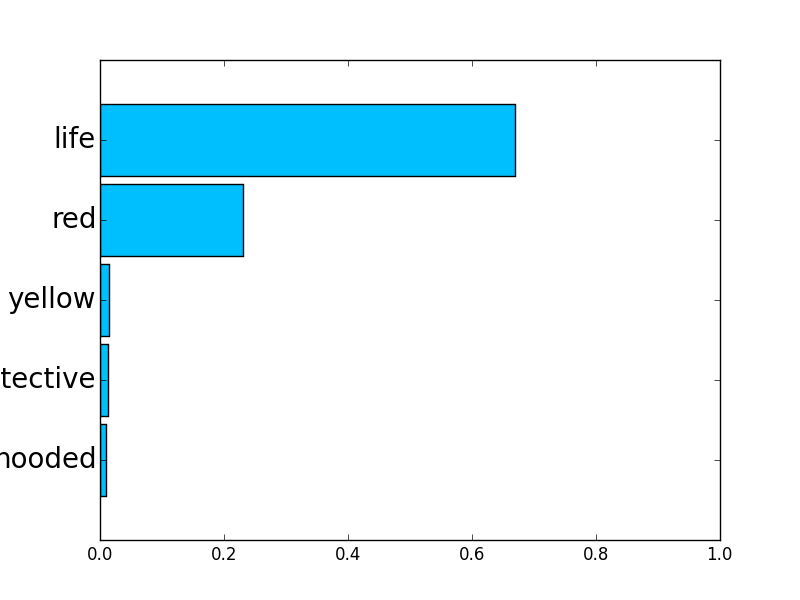}
\end{minipage}
\\

3&\begin{minipage}{0.20\textwidth}
\includegraphics[width=0.70\linewidth]{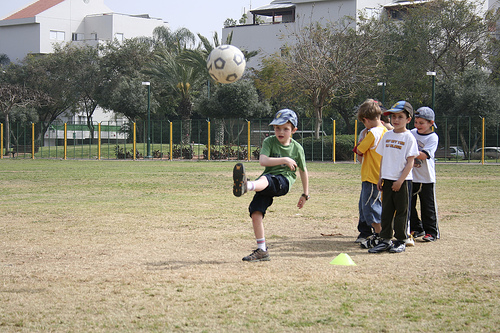}
\end{minipage} 
&
 \begin{minipage}{0.20\textwidth}
\small{ a little boy in a green shirt \textcolor{red}{\textbf{kicks}} a ball}  
\end{minipage}
& 
\begin{minipage}{0.20\textwidth}
\includegraphics[width=0.7\linewidth]{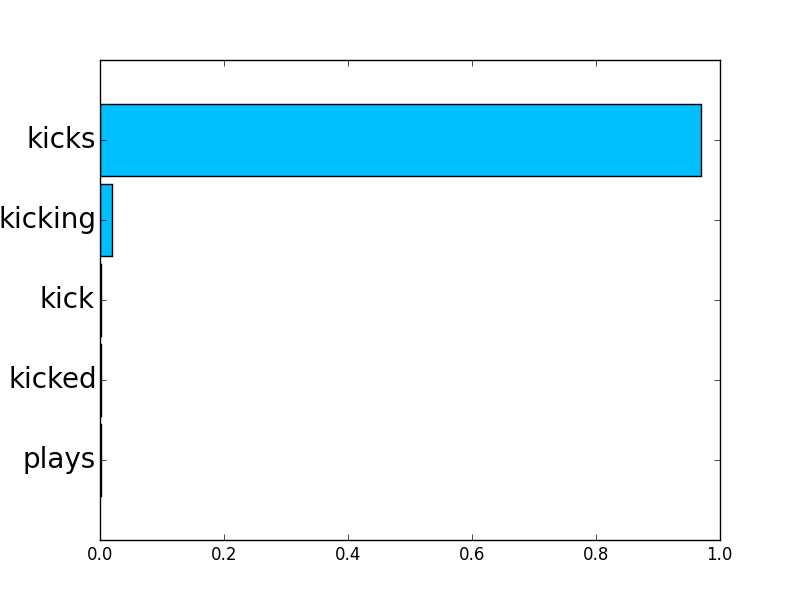}
\end{minipage} 
& 
\begin{minipage}{0.20\textwidth}
\includegraphics[width=0.7\linewidth]{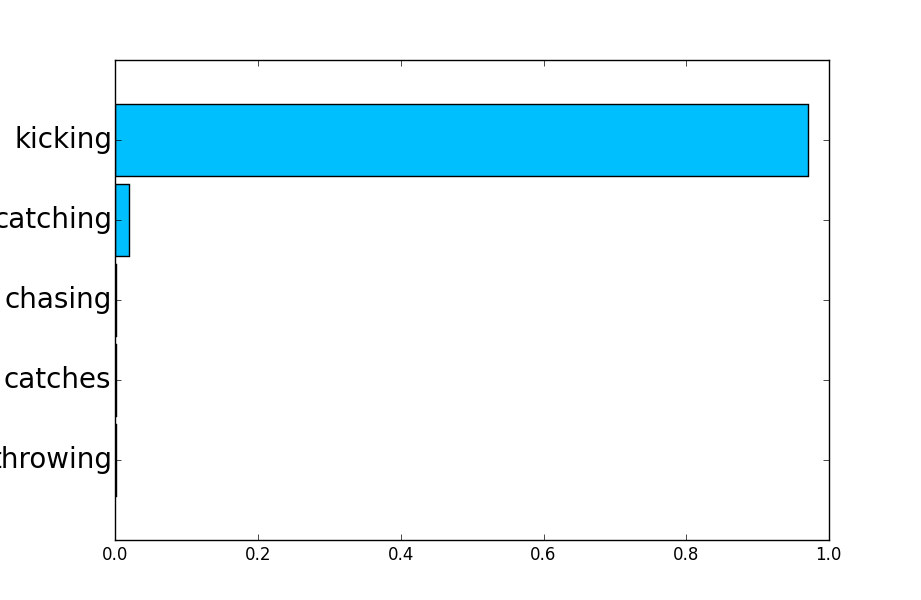}
\end{minipage}
\\

\bottomrule[0.8pt]
\end{tabular}
\caption{Examples of cloze test predictions for ERNIE-ViL pre-training with and without \textbf{SGP} tasks. Masked token are colored in bold and red. The probabilities of the top 5 predictions, denoted as the light purple bars, are listed in the right columns.}
\label{tab_cloze}
\end{table*}

We compare ERNIE-ViL against other cross-modal pre-training models and the results are illustrated in Table \ref{tab_result}.

\begin{table}[t]
  \centering
  \renewcommand{\arraystretch}{1.2}
  \tabcolsep 6.4pt
   \small
 \begin{tabular}{llllllllllll}

\toprule[0.7pt]
\multirow{2}{*}{Nodes} & &
\multicolumn{2}{c}{without \textbf{SGP} tasks}&
&
\multicolumn{2}{c}{with \textbf{SGP} tasks} &

\\
\cline{3-4}
\cline{6-7}
 & \multicolumn{1}{c}{} & \multicolumn{1}{c}{\texttt{ACC@1}} & \multicolumn{1}{c}{\texttt{ACC@5}} &
 &
 \multicolumn{1}{c}{\texttt{ACC@1}} &
 \multicolumn{1}{c}{\texttt{ACC@5}} &
 \\
\hline
objects & \multicolumn{1}{c}{} & \multicolumn{1}{c}{57.14} & \multicolumn{1}{c}{79.22}  & \multicolumn{1}{c}{} & \multicolumn{1}{c}{\textbf{58.34}} & \multicolumn{1}{c}{\textbf{80.80}} \\
attributes & \multicolumn{1}{c}{} & \multicolumn{1}{c}{44.32} & \multicolumn{1}{c}{67.58}  & \multicolumn{1}{c}{} & \multicolumn{1}{c}{\textbf{46.16}} & \multicolumn{1}{c}{\textbf{70.30}} \\
relationships & \multicolumn{1}{c}{} & \multicolumn{1}{c}{47.57} & \multicolumn{1}{c}{68.10}  & \multicolumn{1}{c}{} & \multicolumn{1}{c}{\textbf{50.65}} & \multicolumn{1}{c}{\textbf{71.54}} \\
overall & \multicolumn{1}{c}{} & \multicolumn{1}{c}{49.75} & \multicolumn{1}{c}{71.75}  & \multicolumn{1}{c}{} & \multicolumn{1}{c}{\textbf{51.75}} & \multicolumn{1}{c}{\textbf{74.31}} \\

\bottomrule[0.7pt]
\end{tabular}
\caption{Cloze test results for ERNIE-ViL. An improvement of 2.0\% on overall ACC@1 between models with/without \textbf{SGP} tasks. }
\label{tab_clone_acc}
\end{table}

Among the methods pre-trained on the same out-of-domain datasets (CC and SBU),  
ERNIE-ViL obtains the best performances on all 5 downstream tasks. For the visual reasoning tasks, ERNIE-ViL-large achieves a significant improvement of 6.60\% on VCR (Q$\to$AR) task and 1.74\% on VQA (test-std) task compared with VLBERT-large. On visual grounding task, ERNIE-ViL-large obtains an improvement of 2.40\% for both testA split and testB split on RefCOCO+ task compared to VLBERT-large. On the cross-modal retrieval tasks, where no large models pre-trained on out-of-domain datasets has released results, ERNIE-ViL-base achieves an imporvement of 2.94\% on R@1 for image retrieval and 0.50\% on R@1 for text retrieval compared with Unicoder-VL-base.

For further comparison with those models pretrained with both out-of-domain and in-domain datasets, we  pre-train ERINE-ViL with all these datasets. As illustrated in Table \ref{tab_result}, ERINE-ViL-large acheives state-of-the-art performances on these tasks compared to existing works, e.g., UNITER, OSCAR\cite{li2020oscar} and VILLA\cite{gan2020large}.

\subsection{Analysis}

\paragraph{Effectiveness of Scene Graph Prediction tasks} 

To verify the effectiveness of Scene Graph Prediction (\textbf{SGP}) tasks, we first conduct experiments with ERNIE-ViL-base settings based on the text parameters  initialized from BERT. As illustrated in Table \ref{tab_abl}, pre-training with \textbf{SGP} tasks
in ERNIE-ViL brings significant improvements across all downstream tasks. Especially on Grounding Referring Expressions and Retrieval tasks, those require understanding detailed semantics alignments, \textbf{SGP} tasks make an improvement of 0.69\% accuracy on RefCOCO+ and 2.22\% of R@1 for image retrieval on Flickr30K.

Note that text parameter initialized from ERNIE 2.0 can  lead to further improvements on all tasks and a relatively large improvement on VCR task. We considere that through continually learning on various pre-training tasks, ERNIE 2.0 learned more common sense knowledge which benefits the VCR task. 

Overall, the \textbf{SGP} tasks significantly contribute to the state-of-the-art results of ERNIE-ViL.

\paragraph{Cloze Test} To get a more intuitively understanding of the improvements brought by SGP tasks,  we conduct the language cloze test conditioned on the visual modality. In the cloze test, language tokens represent detailed semantics (objects, attributes and relationships) are masked from the text and the model is required to infer them with the context from both the text and the image. To construct the dataset, we sampled 15,000 image-text pairs from Flickr30K dataset and 5,000 objects, attributes and relationships tokens 
each are selected. For the prediction, the top one accuracy (ACC@1) and top five accuracy (ACC@5) are adopted as the evaluation metric. The comparison of  prediction results between two models, which are pre-trained models with \textbf{SGP} task and without \textbf{SGP} task, are illustrated in Table \ref{tab_clone_acc}. 
The text-stream parameters of both models are initialized from BERT. An absolute improvement of 1.20\% for objects, 3.08\% for relationships and 1.84\% for attributes on ACC@1 demonstrates that ERNIE-ViL pre-trained with \textbf{SGP} tasks learns better cross-modal detailed semantics alignments. 

Moreover, we illustrate some cases in Table \ref{tab_cloze}, and the top $5$ possible predictions are shown in the right columns. As in case 1-2,  model pre-trained without \textbf{SGP} tasks cannot make the right predictions as it didn't learn accurate alignments of detailed semantics, without distinguishing common words and detailed semantics words while pre-training.
While in case 3, the model can predict the reasonable tokens but with lower confidence compared with model pre-trained with \textbf{SGP} tasks.

\section{Conclusion}

We proposed ERNIE-ViL to learn the joint representations of vision and language. In addition to conventional MLM for cross-modal pre-training, we introduce Scene graph Prediction tasks to characterize the cross-modal detailed semantic alignments. Experiment results on various downstream tasks demonstrate the improvements of incorporating structured knowledge obtained from scene graphs during cross-modal pre-training. For future work, scene graphs extracted from images could also be incorporated into cross-modal pre-training. Moreover, Graph Neural Networks that integrate more structured knowledge could be considered as well.

\bibliography{references}

\begin{thebibliography}{31}
\providecommand{\natexlab}[1]{#1}
\providecommand{\url}[1]{\texttt{#1}}
\providecommand{\urlprefix}{URL }
\expandafter\ifx\csname urlstyle\endcsname\relax
  \providecommand{\doi}[1]{doi:\discretionary{}{}{}#1}\else
  \providecommand{\doi}{doi:\discretionary{}{}{}\begingroup
  \urlstyle{rm}\Url}\fi

\bibitem[{Anderson et~al.(2016)Anderson, Fernando, Johnson, and
  Gould}]{anderson2016spice}
Anderson, P.; Fernando, B.; Johnson, M.; and Gould, S. 2016.
\newblock Spice: Semantic propositional image caption evaluation.
\newblock In \emph{European Conference on Computer Vision}, 382--398. Springer.

\bibitem[{Anderson et~al.(2018)Anderson, He, Buehler, Teney, Johnson, Gould,
  and Zhang}]{anderson2018bottom}
Anderson, P.; He, X.; Buehler, C.; Teney, D.; Johnson, M.; Gould, S.; and
  Zhang, L. 2018.
\newblock Bottom-up and top-down attention for image captioning and visual
  question answering.
\newblock In \emph{Proceedings of the IEEE conference on computer vision and
  pattern recognition}, 6077--6086.

\bibitem[{Antol et~al.(2015)Antol, Agrawal, Lu, Mitchell, Batra,
  Lawrence~Zitnick, and Parikh}]{antol2015vqa}
Antol, S.; Agrawal, A.; Lu, J.; Mitchell, M.; Batra, D.; Lawrence~Zitnick, C.;
  and Parikh, D. 2015.
\newblock Vqa: Visual question answering.
\newblock In \emph{Proceedings of the IEEE international conference on computer
  vision}, 2425--2433.

\bibitem[{Chen et~al.(2019)Chen, Li, Yu, Kholy, Ahmed, Gan, Cheng, and
  Liu}]{chen2019uniter}
Chen, Y.-C.; Li, L.; Yu, L.; Kholy, A.~E.; Ahmed, F.; Gan, Z.; Cheng, Y.; and
  Liu, J. 2019.
\newblock Uniter: Learning universal image-text representations.
\newblock \emph{arXiv preprint arXiv:1909.11740} .

\bibitem[{Devlin et~al.(2018)Devlin, Chang, Lee, and
  Toutanova}]{devlin2018bert}
Devlin, J.; Chang, M.-W.; Lee, K.; and Toutanova, K. 2018.
\newblock Bert: Pre-training of deep bidirectional transformers for language
  understanding.
\newblock \emph{arXiv preprint arXiv:1810.04805} .

\bibitem[{Gan et~al.(2020)Gan, Chen, Li, Zhu, Cheng, and Liu}]{gan2020large}
Gan, Z.; Chen, Y.-C.; Li, L.; Zhu, C.; Cheng, Y.; and Liu, J. 2020.
\newblock Large-Scale Adversarial Training for Vision-and-Language
  Representation Learning.
\newblock \emph{arXiv preprint arXiv:2006.06195} .

\bibitem[{Huang et~al.(2020)Huang, Zeng, Liu, Fu, and Fu}]{huang2020pixel}
Huang, Z.; Zeng, Z.; Liu, B.; Fu, D.; and Fu, J. 2020.
\newblock Pixel-BERT: Aligning Image Pixels with Text by Deep Multi-Modal
  Transformers.
\newblock \emph{arXiv preprint arXiv:2004.00849} .

\bibitem[{Johnson, Gupta, and Fei-Fei(2018)}]{johnson2018image}
Johnson, J.; Gupta, A.; and Fei-Fei, L. 2018.
\newblock Image generation from scene graphs.
\newblock In \emph{Proceedings of the IEEE conference on computer vision and
  pattern recognition}, 1219--1228.

\bibitem[{Johnson et~al.(2015)Johnson, Krishna, Stark, Li, Shamma, Bernstein,
  and Fei-Fei}]{johnson2015image}
Johnson, J.; Krishna, R.; Stark, M.; Li, L.-J.; Shamma, D.; Bernstein, M.; and
  Fei-Fei, L. 2015.
\newblock Image retrieval using scene graphs.
\newblock In \emph{Proceedings of the IEEE conference on computer vision and
  pattern recognition}, 3668--3678.

\bibitem[{Kazemzadeh et~al.(2014)Kazemzadeh, Ordonez, Matten, and
  Berg}]{kazemzadeh2014referitgame}
Kazemzadeh, S.; Ordonez, V.; Matten, M.; and Berg, T. 2014.
\newblock Referitgame: Referring to objects in photographs of natural scenes.
\newblock In \emph{Proceedings of the 2014 conference on empirical methods in
  natural language processing (EMNLP)}, 787--798.

\bibitem[{Krishna et~al.(2017)Krishna, Zhu, Groth, Johnson, Hata, Kravitz,
  Chen, Kalantidis, Li, Shamma et~al.}]{krishna2017visual}
Krishna, R.; Zhu, Y.; Groth, O.; Johnson, J.; Hata, K.; Kravitz, J.; Chen, S.;
  Kalantidis, Y.; Li, L.-J.; Shamma, D.~A.; et~al. 2017.
\newblock Visual genome: Connecting language and vision using crowdsourced
  dense image annotations.
\newblock \emph{International Journal of Computer Vision} 123(1): 32--73.

\bibitem[{Li et~al.(2019{\natexlab{a}})Li, Duan, Fang, Jiang, and
  Zhou}]{li2019unicoder}
Li, G.; Duan, N.; Fang, Y.; Jiang, D.; and Zhou, M. 2019{\natexlab{a}}.
\newblock Unicoder-vl: A universal encoder for vision and language by
  cross-modal pre-training.
\newblock \emph{arXiv preprint arXiv:1908.06066} .

\bibitem[{Li et~al.(2019{\natexlab{b}})Li, Yatskar, Yin, Hsieh, and
  Chang}]{li2019visualbert}
Li, L.~H.; Yatskar, M.; Yin, D.; Hsieh, C.-J.; and Chang, K.-W.
  2019{\natexlab{b}}.
\newblock Visualbert: A simple and performant baseline for vision and language.
\newblock \emph{arXiv preprint arXiv:1908.03557} .

\bibitem[{Li et~al.(2020)Li, Yin, Li, Hu, Zhang, Zhang, Wang, Hu, Dong, Wei
  et~al.}]{li2020oscar}
Li, X.; Yin, X.; Li, C.; Hu, X.; Zhang, P.; Zhang, L.; Wang, L.; Hu, H.; Dong,
  L.; Wei, F.; et~al. 2020.
\newblock Oscar: Object-semantics aligned pre-training for vision-language
  tasks.
\newblock \emph{arXiv preprint arXiv:2004.06165} .

\bibitem[{Lin et~al.(2014)Lin, Maire, Belongie, Hays, Perona, Ramanan,
  Doll{\'a}r, and Zitnick}]{lin2014microsoft}
Lin, T.-Y.; Maire, M.; Belongie, S.; Hays, J.; Perona, P.; Ramanan, D.;
  Doll{\'a}r, P.; and Zitnick, C.~L. 2014.
\newblock Microsoft coco: Common objects in context.
\newblock In \emph{European conference on computer vision}, 740--755. Springer.

\bibitem[{Lu et~al.(2019)Lu, Batra, Parikh, and Lee}]{lu2019vilbert}
Lu, J.; Batra, D.; Parikh, D.; and Lee, S. 2019.
\newblock Vilbert: Pretraining task-agnostic visiolinguistic representations
  for vision-and-language tasks.
\newblock In \emph{Advances in Neural Information Processing Systems}, 13--23.

\bibitem[{Ordonez, Kulkarni, and Berg(2011)}]{ordonez2011im2text}
Ordonez, V.; Kulkarni, G.; and Berg, T.~L. 2011.
\newblock Im2text: Describing images using 1 million captioned photographs.
\newblock In \emph{Advances in neural information processing systems},
  1143--1151.

\bibitem[{Radford et~al.(2018)Radford, Narasimhan, Salimans, and
  Sutskever}]{radford2018improving}
Radford, A.; Narasimhan, K.; Salimans, T.; and Sutskever, I. 2018.
\newblock Improving language understanding by generative pre-training.
\newblock In \emph{URL
  https://s3-us-west-2.amazonaws.com/openaiassets/research-covers/language
  unsupervised/language understanding paper.pdf}.

\bibitem[{Sharma et~al.(2018)Sharma, Ding, Goodman, and
  Soricut}]{sharma2018conceptual}
Sharma, P.; Ding, N.; Goodman, S.; and Soricut, R. 2018.
\newblock Conceptual captions: A cleaned, hypernymed, image alt-text dataset
  for automatic image captioning.
\newblock In \emph{Proceedings of the 56th Annual Meeting of the Association
  for Computational Linguistics (Volume 1: Long Papers)}, 2556--2565.

\bibitem[{Su et~al.(2019)Su, Zhu, Cao, Li, Lu, Wei, and Dai}]{su2019vl}
Su, W.; Zhu, X.; Cao, Y.; Li, B.; Lu, L.; Wei, F.; and Dai, J. 2019.
\newblock Vl-bert: Pre-training of generic visual-linguistic representations.
\newblock \emph{arXiv preprint arXiv:1908.08530} .

\bibitem[{Sun et~al.(2020)Sun, Cheng, Zhang, Zhang, Zheng, Wang, and
  Wei}]{sun2020circle}
Sun, Y.; Cheng, C.; Zhang, Y.; Zhang, C.; Zheng, L.; Wang, Z.; and Wei, Y.
  2020.
\newblock Circle loss: A unified perspective of pair similarity optimization.
\newblock \emph{arXiv preprint arXiv:2002.10857} .

\bibitem[{Sun et~al.(2019)Sun, Wang, Li, Feng, Tian, Wu, and
  Wang}]{sun2019ernie2}
Sun, Y.; Wang, S.; Li, Y.; Feng, S.; Tian, H.; Wu, H.; and Wang, H. 2019.
\newblock Ernie 2.0: A continual pre-training framework for language
  understanding.
\newblock \emph{arXiv preprint arXiv:1907.12412} .

\bibitem[{Tan and Bansal(2019)}]{tan2019lxmert}
Tan, H.; and Bansal, M. 2019.
\newblock Lxmert: Learning cross-modality encoder representations from
  transformers.
\newblock \emph{arXiv preprint arXiv:1908.07490} .

\bibitem[{Vaswani et~al.(2017)Vaswani, Shazeer, Parmar, Uszkoreit, Jones,
  Gomez, Kaiser, and Polosukhin}]{vaswani2017attention}
Vaswani, A.; Shazeer, N.; Parmar, N.; Uszkoreit, J.; Jones, L.; Gomez, A.~N.;
  Kaiser, {\L}.; and Polosukhin, I. 2017.
\newblock Attention is all you need.
\newblock In \emph{Advances in neural information processing systems},
  5998--6008.

\bibitem[{Wu et~al.(2019)Wu, Mao, Zhang, Jiang, Li, Sun, and
  Ma}]{wu2019unified}
Wu, H.; Mao, J.; Zhang, Y.; Jiang, Y.; Li, L.; Sun, W.; and Ma, W.-Y. 2019.
\newblock Unified visual-semantic embeddings: Bridging vision and language with
  structured meaning representations.
\newblock In \emph{Proceedings of the IEEE Conference on Computer Vision and
  Pattern Recognition}, 6609--6618.

\bibitem[{Yang et~al.(2019)Yang, Tang, Zhang, and Cai}]{yang2019auto}
Yang, X.; Tang, K.; Zhang, H.; and Cai, J. 2019.
\newblock Auto-encoding scene graphs for image captioning.
\newblock In \emph{Proceedings of the IEEE Conference on Computer Vision and
  Pattern Recognition}, 10685--10694.

\bibitem[{Young et~al.(2014)Young, Lai, Hodosh, and
  Hockenmaier}]{young2014image}
Young, P.; Lai, A.; Hodosh, M.; and Hockenmaier, J. 2014.
\newblock From image descriptions to visual denotations: New similarity metrics
  for semantic inference over event descriptions.
\newblock \emph{Transactions of the Association for Computational Linguistics}
  2: 67--78.

\bibitem[{Yu et~al.(2018)Yu, Lin, Shen, Yang, Lu, Bansal, and
  Berg}]{yu2018mattnet}
Yu, L.; Lin, Z.; Shen, X.; Yang, J.; Lu, X.; Bansal, M.; and Berg, T.~L. 2018.
\newblock Mattnet: Modular attention network for referring expression
  comprehension.
\newblock In \emph{Proceedings of the IEEE Conference on Computer Vision and
  Pattern Recognition}, 1307--1315.

\bibitem[{Zellers et~al.(2019)Zellers, Bisk, Farhadi, and
  Choi}]{zellers2019recognition}
Zellers, R.; Bisk, Y.; Farhadi, A.; and Choi, Y. 2019.
\newblock From recognition to cognition: Visual commonsense reasoning.
\newblock In \emph{Proceedings of the IEEE Conference on Computer Vision and
  Pattern Recognition}, 6720--6731.

\bibitem[{Zhang, Chao, and Xuan(2019)}]{zhang2019empirical}
Zhang, C.; Chao, W.-L.; and Xuan, D. 2019.
\newblock An empirical study on leveraging scene graphs for visual question
  answering.
\newblock \emph{arXiv preprint arXiv:1907.12133} .

\bibitem[{Zhou et~al.(2019)Zhou, Palangi, Zhang, Hu, Corso, and
  Gao}]{zhou2019unified}
Zhou, L.; Palangi, H.; Zhang, L.; Hu, H.; Corso, J.~J.; and Gao, J. 2019.
\newblock Unified vision-language pre-training for image captioning and vqa.
\newblock \emph{arXiv preprint arXiv:1909.11059} .

\end{thebibliography}

\end{document}